\documentclass[runningheads]{llncs}
\usepackage[T1]{fontenc}
\usepackage{graphicx}
\usepackage{color}
\usepackage{graphicx}
\usepackage{booktabs}
\usepackage{accvabbrv}

\usepackage{algorithm,xcolor}
\usepackage{algpseudocode}
\usepackage{amsfonts}
\usepackage{rotating}
\usepackage{import}
\begin{document}
\title{A Meaningful Perturbation Metric for Evaluating Explainability Methods}
\author{Danielle Cohen \and
Hila Chefer \and
Lior Wolf}
\authorrunning{D. Cohen et al.}
\institute{Blavatnik School of Computer Science, Tel Aviv University}
\maketitle              % typeset the header of the contribution
\begin{abstract}
  Deep neural networks (DNNs) have demonstrated remarkable success, yet their wide adoption is often hindered by their opaque decision-making. To address this, attribution methods have been proposed to assign relevance values to each part of the input. However, different methods often produce entirely different relevance maps, necessitating the development of standardized metrics to evaluate them. Typically, such evaluation is performed through perturbation, wherein high- or low-relevance regions of the input image are manipulated to examine the change in  prediction. In this work, we introduce a novel approach, which harnesses image generation models to perform targeted perturbation. Specifically, we focus on inpainting only the high-relevance pixels of an input image to modify the model's predictions while preserving image fidelity. This is in contrast to existing approaches, which often produce out-of-distribution modifications, leading to unreliable results. Through extensive experiments, we demonstrate the effectiveness of our approach in generating meaningful rankings across a wide range of models and attribution methods. Crucially, we establish that the ranking produced by our metric exhibits significantly higher correlation with human preferences compared to existing approaches, underscoring its potential for enhancing interpretability in DNNs.

\end{abstract}
\section{Introduction}
\label{sec:intro}
In recent years, deep neural networks (DNNs) have revolutionized research in countless domains. Yet, unlike some predecessors \eg, decision trees and linear regressors, DNN predictions remain opaque. DNNs often rely on spurious correlations or ``shortcuts'', leading to biases~\cite{Buolamwini2018GenderSI,Fuster2017PredictablyUT} or robustness issues~\cite{Hendrycks2019BenchmarkingNN,Hendrycks2019NaturalAE,chefer2022robustvit} that hamper their widespread adoption. 

% ~\cite{Hendrycks2019BenchmarkingNN,Hendrycks2020TheMF,Hendrycks2019NaturalAE,chefer2022robustvit}

%
%including Natural Language %Processing~\cite{Vaswani2017AttentionIA,Radford2018ImprovingLU}, Computer %Vision~\cite{dosovitskiy2020image,Rombach2021HighResolutionIS}, %medicine~\cite{Rav2017DeepLF,Miotto2016DeepPA}, and more. 
%

To overcome this obstacle, several explainability and interpretability techniques have been proposed to clarify the decision-making process of DNNs~\cite{selvaraju2017grad,smilkov2017SmoothGrad,chen2018lshapley,Chefer_2021_CVPR,Goyal2019CounterfactualVE}. In this work, we focus on attribution methods, which constitute one of the most prominent approaches for producing such interpretations. Attribution methods assign a corresponding relevance value to each of the input parts (e.g., image pixels), where a higher relevance score suggests a greater impact on the model's prediction. Various attribution methods have been proposed to interpret the predictions by DNNs, each claiming to be the most ``faithful'' to the model, while often producing entirely different results than their counterparts. However, in the absence of a rigorous definition of faithfulness and a set of standard acceptable evaluation metrics, it becomes impossible to decipher which attribution method should be employed for a given use case.

\begin{figure}[t!]
    % \centering
    % \setlength{\tabcolsep}{0.5pt}
    % \addtolength{\belowcaptionskip}{-8pt}
    % {\small
    % \begin{tabular}{c c c c}
        
    %     {\includegraphics[scale=0.255]{images/rand_10_perc/resnet_original_class_1(goldfish, Carassius auratus).png}} \hspace{0.05cm} &
    %     {\includegraphics[scale=0.255]{images/rand_10_perc/resnet_original_class_1_rand_0.1_class_980(volcano).png}} \hspace{0.05cm}
    %     & 
    %     {\includegraphics[scale=0.255]{images/rand_10_perc/vit_original_class_914(yawl).png}} \hspace{0.05cm} 
    %     &
    %     {\includegraphics[scale=0.255]{images/rand_10_perc/vit_original_class_914_rand_0.1_class_611(jigsaw puzzle).png}} \hspace{0.05cm}\\
    %     Goldfish & Volcano & Yawl & Jigsaw puzzle \\
    %     \multicolumn{2}{c}{(a)} & \multicolumn{2}{c}{(b)}

    % \end{tabular}
    % }
     \begin{center}
             \includegraphics[width=0.82\linewidth]{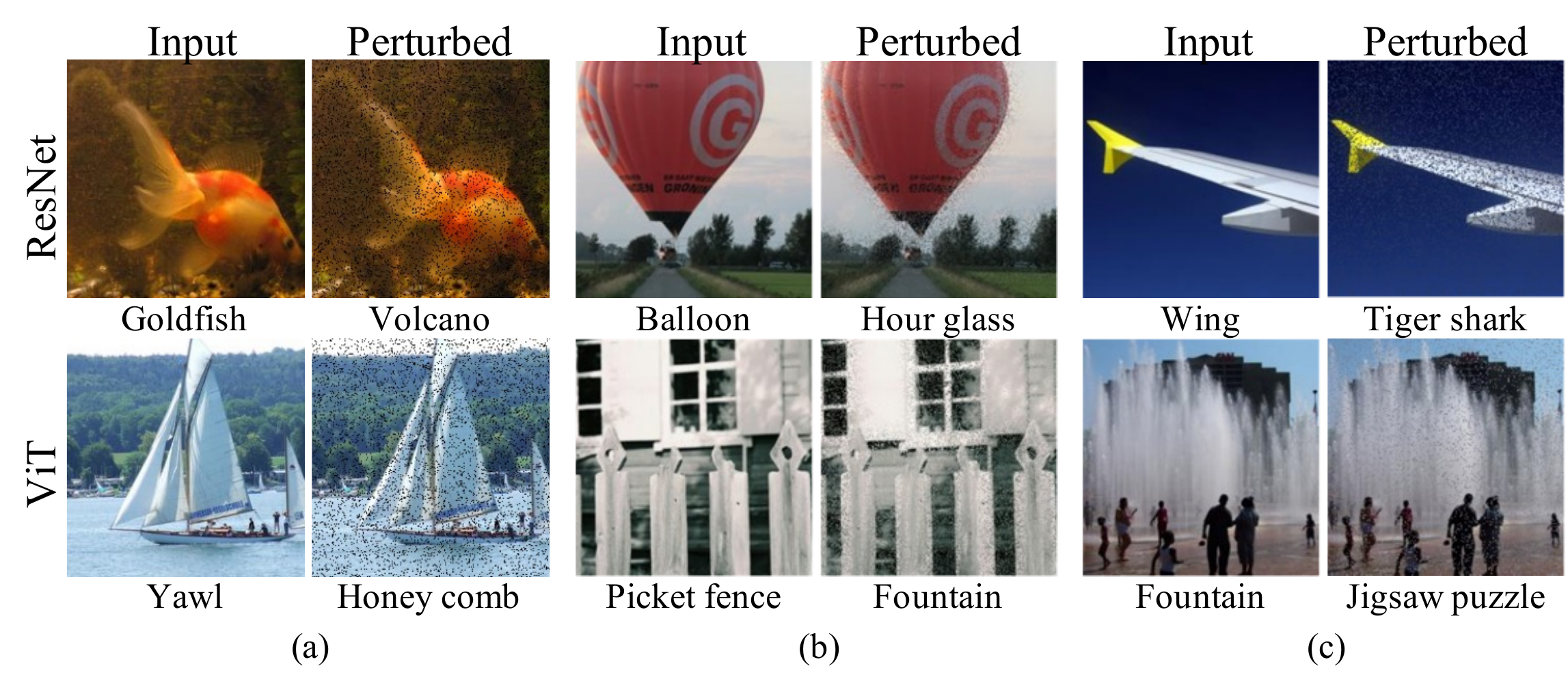}
        \end{center}
    \vspace{-20px}
    \caption{\textbf{Examples of OOD predictions following perturbation} with (a) pixel deletion \cite{hooker2019benchmark} (b) pixel blurring \cite{fong2017interpretable} and (c) per channel mean \cite{hooker2019benchmark} on both ResNet \cite{He2015DeepRL} and ViT \cite{dosovitskiy2020image} backbones. For each image, we randomly draw pixels and apply the perturbation method. We enclose the original and modified prediction. As can be seen, even when the majority of the content is still visible, the model is sensitive to the OOD effect of the perturbation, causing it to modify its prediction.  
    }
    \label{fig:rand_mask}
    \vspace{-18px}
\end{figure}

In this work we adopt the dominant notion of faithfulness, where a relevance map is considered to be ``faithful'' to the model if it reflects the true reasoning process behind the model's prediction~\cite{jacovi-goldberg-2020-towards,Liu2022RethinkingAE,Chefer_2021_CVPR}. In other words, given a relevance map, our metric should estimate the correlation between the pixels with high relevance values and the prediction made by the model. Another important parameter for the evaluation of attribution methods is ``plausibility'', \ie, how convincing the explanation is to humans~\cite{jacovi-goldberg-2020-towards}. Both faithfulness and plausibility are important when evaluating an explanation. Without faithfulness, there is no direct connection between the relevance map and the model we attempt to explain, and therefore the explanation has no utility~\cite{adebayo2018sanity}. Without plausibility, the explanation may not be clear to humans, which defies its goal of helping humans understand the reasoning behind the decision-making~\cite{Fel2021WhatIC}.

The predominant approach to evaluating attribution maps in the context of image classifiers is through image perturbation~\cite{Li2016UnderstandingNN,hooker2019benchmark,DeYoung2019ERASERAB,fong2017interpretable}. The intuition is that pixels are perturbed from most (least) relevant. If high (low) relevance marks important (unimportant) pixels, perturbing alters (keeps) the prediction.

% The intuition behind perturbation is simple; pixels are gradually perturbed from most (least) relevant. Assuming that high (low) relevance values are assigned to important (unimportant) pixels, perturbing the high (low) relevance pixels will cause the prediction to change (remain the same). 

However, existing perturbation methods make unnatural modifications to the input image, such as deleting pixels~\cite{hooker2019benchmark,Liu2022RethinkingAE}, blurring pixels~\cite{fong2017interpretable} or per channel mean~\cite{hooker2019benchmark}. As we demonstrate in Fig.~\ref{fig:rand_mask} these alterations often drive the input out of distribution, leading the model to produce predictions that do not correlate with the image pixels. Importantly, in Fig.~\ref{fig:rand_mask} we apply \emph{completely random masks} such that the content of the image is still mostly preserved, and even so, the perturbation causes the model to change its prediction. This random baseline demonstrates the significant challenge to disentangling between the OOD effect of the perturbation approach on the prediction and the relevance of the modified pixels to the prediction, making existing perturbation approaches unreliable.

Our research hypothesis is that a better approach is to use the perturbed pixels in a way that presents the classifier with a valid alternative. The easiest alternative target is likely to be the class with the second-highest classification score. We employ the capabilities of Stable Diffusion (SD) inpainting~\cite{Rombach2021HighResolutionIS} with the appropriate conditioning to modify the high relevance pixels, such that they will align with the second highest prediction, and test whether the prediction changes to \emph{the specified target class}. This allows us to filter out modifications that lead to unrelated classes, such as the ones in Fig.~\ref{fig:rand_mask}. Additionally, we add a novel weighting function that compares the relevance maps before and after the inpainting procedure. This validates that the change in prediction was indeed \emph{caused} by the inpainting procedure, further establishing the connection between the perturbed pixels and the modification in the prediction.

We perform extensive experiments to compare our Stratified Inpainting approach to existing perturbation approaches, such as deleting pixels, blurring pixels, and per channel mean. We test different prominent attribution approaches~\cite{selvaraju2017grad,smilkov2017SmoothGrad,abnar2020quantifying,chen2018lshapley,Chefer_2021_CVPR}, and various image backbones, such as different variants of CNNs~\cite{He2015DeepRL,Simonyan2014VeryDC} and Vision Transformers~\cite{dosovitskiy2020image}, and find that in many of the cases, existing methods struggle to provide a meaningful ranking and that the perturbation graphs do not give a meaningful distinction between methods. Surprisingly, we show that the existing perturbation methods often struggle to separate between valid attribution methods and \emph{randomly generated relevance maps}, further indicating the inadequacy of employing these methods for evaluating explanations. Conversely, our Stratified Inpainting approach maintains decent separation between methods and produces robust results across different models and datasets. Via a user study, we find that the ranking produced by our metric is much more correlated with human preferences, demonstrating that in the case of attribution methods, faithfulness and plausibility are not mutually exclusive, and are both identified by our metric simultaneously.
{\section{Related Work}

\noindent{\bf Explainability in computer vision.~} Many methods were proposed for generating heatmaps to explain deep networks. {\em Gradient based} methods use loss gradients via backpropagation \cite{shrikumar2017learning,Sundararajan2017AxiomaticAF,smilkov2017SmoothGrad,srinivas2019full,selvaraju2017grad}. {\em Attribution propagation} methods, such as LRP, recursively decompose decisions based on the Deep Taylor Decomposition framework \cite{bach2015pixel,montavon2017explaining,binder2016layer}. Other methods that do not fall into these two categories include saliency-based methods~\cite{dabkowski2017real,simonyan2013deep,mahendran2016visualizing,zhou2016learning,zeiler2014visualizing}, {Activation Maximization}~\cite{erhan2009visualizing}, Excitation Backprop~\cite{zhang2018top}, concept-based methods~\cite{Kim2017InterpretabilityBF,Ghorbani2019TowardsAC}, and Perturbation methods~\cite{fong2019understanding,fong2017interpretable,lundberg2017unified,chen2018lshapley}, which analyze network responses to input changes. With the rise of Transformers, attention explainability methods have gained interest. LRP was applied to Transformers \cite{voita2019analyzing}, while methods like rollout \cite{abnar2020quantifying} and Transformer attribution \cite{Chefer_2021_CVPR,Chefer_2021_ICCV} extend LRP and gradients to explain attention mechanisms.

\paragraph{\bf Generative models for explainability.~} Several works have investigated the use of generative models for explainability-related tasks. Agarwal et al.\cite{agarwal2020explaining} and Chang et al. \cite{chang2018explaining} proposed incorporating inpaiting with existing attribution methods \cite{fong2017interpretable} to improve quality. While sharing a similar intuition, we introduce a metric for evaluating existing attribution methods rather than proposing a new one. Critically, fully inpainting-based attribution can be computationally expensive necessitates at least 300 iterations \cite{agarwal2020explaining}. In contrast, by leveraging inpainting only for evaluation, our metric reduces this overhead to just 10 iterations, greatly easing the computational burden. Finally, note that our metric also proposes a novel data filtering method, and a weighting function to directly connect the inpainted pixels to the modification in the classification.

% \smallskip
\paragraph{\bf Evaluating explanations.~} While the research around explainability algorithms continues to evolve, developing methods to evaluate the provided explanations remains a less explored field of research. Most metrics of evaluating explanations are based on perturbation tests and segmentation results, which can be seen as a general case of the Pointing-Game~\cite{hooker2019benchmark,fong2017interpretable}. More recent works \cite{Liu2022RethinkingAE} have suggested testing the confidence gap induced by removing features to evaluate the faithfulness of the explanation to the explained model. A different line of work \cite{Fel2021WhatIC} evaluates the explanations based on their utility to humans, i.e., the information humans can deduce from the explanations. Another related area is counterfactual explanations. Goyal et al. \cite{Goyal2019CounterfactualVE} used inpainting to generate counterfactual explanations. The primary distinction between our paper and theirs lies in the task. Our objective is to evaluate attribution methods, we use inpaint to assess the method's attribution maps while counterfactual explanations task is identifying the minimal set of pixels that will change the classification.

% While, as described above, faithfulness can be defined in multiple ways, depending on the objective of the method (human utility, correspondence to the model, etc.), we choose to focus on faithfulness with respect to the explained classifier, i.e., we wish to verify that high relevance is assigned by the method to the pixels that most influence the classifier's prediction, and vice versa.

\paragraph{\bf Text-to-image generation.~} Recent text-to-image generative models have demonstrated an unparalleled ability to generate diverse and creative imagery guided by a target text prompt~\cite{Ramesh2022Hierarchical,nichol2021glide,Rombach2021HighResolutionIS,Saharia2022PhotorealisticTD}. In this work we leverage the powerful Stable Diffusion model~\cite{Rombach2021HighResolutionIS} to produce natural perturbations to images based on text. Specifically, we leverage the Stable Diffusion inpainting model\footnote{The SD inpainting model was trained by Runway ML, and implemented over the diffusers library: \url{https://huggingface.co/runwayml/stable-diffusion-inpainting}}, which was fine-tuned on the inpainting task. Given an image, a mask, and a target text prompt, the model completes the masked areas in a manner that corresponds to the input text prompt.  

\section{Method}
We begin by describing the problem setting. 
We then provide motivation for our method by highlighting the pitfalls of the widely used perturbation metrics and describe our method in detail. 

\subsection{Problem Setting}
Given a classifier $C$ and an explainability method $E$, the goal of our method is to evaluate the faithfulness of the explanations produced by $E$ to the predictions made by $C$.
Formally, given a set of test images to examine, $I=\{i_1, \dots, i_N\}$ s.t. $\forall i\in I: i\in \mathbb{R}^{h,w}$, the explainability algorithm produces a relevance map $E(C, i) = R_{E,C}(i)\in \mathbb{R}^{h,w}$. Our goal is to produce a score $f(C, E, I)$ that reflects the faithfulness of the relevance maps $R_{E,C}(i)$ to the classifier's predictions $C(i)$ on the test set, where a higher score indicates that the maps produced by $E$ are more faithful to the classifier $C$.

\subsection{Stratified Inpainting}
\label{sec:method}
The intuition behind our method is to use the powerful image generation prior of text-to-image diffusion models \cite{Rombach2021HighResolutionIS} to \emph{replace the removed pixels by other plausible pixels} instead of unnaturally perturbing them (\eg, by deletion or blurring), in order to avoid OOD inputs. Additionally, rather than measuring an arbitrary change to any other class (which may be unintuitive, as demonstrated in Fig. \ref{fig:rand_mask}), we assess the relevance's ability to transform the prediction to the \emph{second highest prediction}. In other words, we estimate whether the highlighted features are indeed those that lead to the top-1 prediction, by substituting them with a plausible alternative and examining the change in prediction against a \emph{specific plausible class}.
Next, we describe the steps of our method in detail.

\paragraph{Step 1: Creating the evaluation dataset $I$.} 

Given a classifier $C$ and an inpainting Stable Diffusion model $SD$, we aim curate a test set of classes to extract our test images from. Importantly, while Stable Diffusion is very expressive, it has been demonstrated \cite{2303.17155} that it fails to capture some classes, especially fine-grained classes in datasets such as ImageNet~\cite{russakovsky2015ImageNet}. Therefore, we automatically curate a set $S$ of \emph{classes} recognizable by $SD$. Our Stratified inpainting algorithm is applied over a test set $I$, containing images from the test set such that the top-2 prediction by the classifier is in the set $S$ defined above, i.e., $top_2\left(C(i)\right) \in S$. This way, we insure $SD$ is capable of inpainting the images given the selection of the top-2 class. Please refer to the supplementary materials for additional implementation details on the construction of the dataset.

\paragraph{Step 2: Evaluation on the test classes.}
\algnewcommand{\algorithmicgoto}{\textbf{Go to}}%
\algnewcommand{\Goto}[1]{\algorithmicgoto~\ref{#1}}%
\begin{algorithm}[t!]
\caption{Stratified inpainting}

\begin{flushleft}
% \vspace{-1px}
\textbf{Input:} A classifier to explain $C$, an explainability method $E$, an image to explain $i\in I$,
and a pre-trained inpainting Stable Diffusion model $SD$.\\
\textbf{Output:} Stratified inpainting scores for $p=\{10\%, \dots, 50\%\}$ of the most relevant image pixels.
\end{flushleft}
\begin{algorithmic}[1]
\State $scores \gets \{\}$
\State $R_{E,C}(i) \gets E(C, i)$
\State $R_{E,C}(i) \gets bilinear(R_{E,C}(i), factor=\frac{1}{16})$
\For{$p \in \{10\%, \dots, 50\%\}$}
  % \If{Positive} 
        \State $masked\_patches \gets top_p(R_{E,C}(i))$
   %     \Else{}
   %     \State $masked\_patches \gets bottom_p(R_{E,C}(i))$
   % \EndIf
    \State $\forall patch\in i: mask[p] \gets 1$
    \State $\forall patch\in masked\_patches: mask[p] \gets 0$
    \State $i' \gets SD(i \odot mask, top\_2\_class)$
    \If{$top_1\left(C(i')\right) = top_2\left(C(i)\right)$} 
        \State $w \gets \frac{|top_p\left(R_{E,C_2} (i')\right) \cap \left(top_p\left(R_{E,C_2} (i)\right)\cup M \right)|}{|top_p\left(R_{E,C_2} (i')\right)|}$
        \State $scores \gets \textbf{concat}(scores , \{w\})$
        \Else{}
        \State $scores \gets \textbf{concat}(scores , \{0\})$
    \EndIf
\EndFor
\State \textbf{Return} $scores$
\end{algorithmic}
\label{alg:method}
% \vspace{-2px}
\end{algorithm}
The method for obtaining scores for each test image $i \in I$ is outlined in Alg.~\ref{alg:method}. We begin by calculating the corresponding relevance map $R_{E,C}(i)$ in line 2. Next (line 3), we apply bilinear downsampling to obtain patch-wise relevance values (patch size 16, following \cite{dosovitskiy2020image}), as SD inpainting struggles with distant single pixels. This limitation is alleviated by the use of patch-wise explanations (see ablation study). We consider perturbation steps $p$ between $10\%$ and $50\%$, as inpainting over $50\%$ often obtains the top-2 class by simply adding another object. For each step $p$ we create a mask that captures the top pixels of that step. We use this mask to inpaint (lines 5-8) the top patches of the image by the relevance map $R_{E,C}$(i) (line 5), using the prompt ``\{top-2 class name\}''. If $R_{E,C}(i)$ indeed faithfully captures the most relevant pixels for the prediction, then prediction of the inpainted image should change to the top-2 class. 

In cases where the prediction indeed changes to be the top-2 class, we wish to verify that the change was caused by the set of top $p$ pixels that were perturbed in the inpainting process. To this end, we compare the relevance maps obtained for the original and the inpainted images. A successful inpainting should lead to high relevance either in the inpainted areas (since those are the areas actually modified to fit the top-2 class) or in areas that were already attributed to the top-2 class. Therefore, we add a weighting function to capture this behavior. Let $M$ be the set of top $p$ pixels modified by the inpainting step, $i$ be the original image, and $i'$ be the unpainted image. Let $C_1, C_2$ be the top-1, and top-2 class of $i$, respectively. By our previous notations, $top_p\left(R_{E,C_2} (i)\right), top_p\left(R_{E,C_2} (i')\right)$ are the top pixels in the relevance maps of the original and inpainted images for the top-2 class. We apply the following weighting function (line 10): $w=\frac{|top_p\left(R_{E,C_2} (i')\right) \cap \left(top_p\left(R_{E,C_2} (i)\right)\cup M \right)|}{|top_p\left(R_{E,C_2} (i')\right)|}$. This  indicates the amount of top $p$ pixels in $i'$'s explanation of label $C_2$ that are also in the inpainting mask $M$ from $i$, or are in the original relevance map for class $C_2$ extracted from  $i$. 

We update the score for $p$ with $w$ if the prediction changed to the target class, which is the original top-2 class, and $0$ otherwise (lines 9-14). The list of all scores (one per percentile p) is returned (line 16). 

This process is repeated for all images in $I$ that follow the selection criterion $top_2\left(C(i)\right) \in S$. 
The final score per step $p$ is obtained by averaging the values for that specific step across all such images $i$. Kindly refer to the supplementary materials for additional implementation details.

\section{Experiments}
%\subsection{Experimental setup}
\label{sec:experimental_setup}
{\textbf{Baselines.}}
% {\color{red} TODO DANIELLE add info on the additional baselines, refer to baselines that do not appear in the main paper, and refer the reader to the supp}
We compare our method against the most common perturbation alternatives. First, with perhaps the most common approach, which is pixel deletion \cite{hooker2019benchmark}, and then with other prominent alternatives, such as blurring \cite{fong2017interpretable} and per channel mean \cite{hooker2019benchmark}. We also consider less related methods, such as faithfulness violation \cite{Liu2022RethinkingAE}, and saliency-based evaluation \cite{agarwal2020explaining,dabkowski2017real}. These alternatives are less competitive and need adaptation for comparison, so the full comparison is in the supplementary materials. Similar to Alg. \ref{alg:method} (lines 9-12), each method scores explanation $R_{E,C}(i)$ differently. Importantly, existing techniques check the change of the prediction to \emph{any other class}, thus they are vulnerable to OOD prediction changes, such as those demonstrated in Fig. \ref{fig:rand_mask}. Conversely, our metric measures the modification to \emph{a specific plausible class}, mitigating this OOD effect.

\smallskip
\noindent{\textbf{Models and dataset.}}
We experiment on widely used image backbones: ResNet-50 \cite{He2015DeepRL}, VGG \cite{simonyan2013deep}, AlexNet \cite{krizhevsky2012imagenet} and ViT-B/16 \cite{dosovitskiy2020image}, to demonstrate our method's ability to provide meaningful evaluations across different selections of backbones, both CNN- and transformer-based. 
We construct the dataset for comparison from the subset of classes $S$ collected as described in the previous section, over the ImageNet validation set \cite{guillaumin2014imagenet} and the binary cat-dog dataset for AlexNet\footnote{Taken from kaggle: \url{https://www.kaggle.com/datasets/tongpython/cat-and-dog}}. 

\smallskip
%
%This results in $|S|=661$ classes for ResNet and VGG, $|S|=704$ classes for ViT, and all classes for the binary classifier. We then sample an image from each class in ImageNet as described in Step 2 of our method (see Sec. \ref{sec:method}).
%
\noindent{\textbf{Explainability methods.}} For each model, we use the most common explainability algorithms and \emph{random} pixel selection as a baseline to assess the metric's robustness to OOD effects, as shown in Fig. \ref{fig:rand_mask}. For CNNs, we consider GradCAM (gc) \cite{selvaraju2017grad}, FullGrad \cite{srinivas2019full}, SmoothGrad \cite{smilkov2017SmoothGrad}, Grad SHAP \cite{lundberg2017unified}, inputXgrad \cite{shrikumar2017learning}, IntegratedGradients\cite{Sundararajan2017AxiomaticAF} and layered GradCam \cite{Dhamdhere2018HowII}. For ViT, we consider GradCAM (gc) \cite{selvaraju2017grad}, Transformer Attribution (transformer) \cite{Chefer_2021_CVPR}, and partial LRP (lrp) \cite{voita2019analyzing}, as well as the vanilla attention (attn last layer), and rollout \cite{abnar2020quantifying}.

\subsection{Qualitative results}

\begin{figure*}[t!]
    \centering
    \setlength{\tabcolsep}{0.5pt}
    \addtolength{\belowcaptionskip}{-8pt}
    {\small
    \begin{tabular}{l c c c c}
    &
        ResNet \cite{He2015DeepRL} & VGG \cite{simonyan2013deep} & AlexNet (binary) \cite{krizhevsky2012imagenet} & ViT \cite{dosovitskiy2020image}\\
        \raisebox{5px}{\begin{turn}{90}~~~ Delete \cite{hooker2019benchmark} \end{turn}} 
    % \begin{tabular}{l c c c}
    % &
    %     ResNet \cite{He2015DeepRL} & AlexNet (binary) \cite{krizhevsky2012imagenet} & ViT \cite{dosovitskiy2020image}\\
    %     \raisebox{5px}{\begin{turn}{90}~~~ Delete \cite{hooker2019benchmark} \end{turn}} 
 & {\includegraphics[scale=0.2]{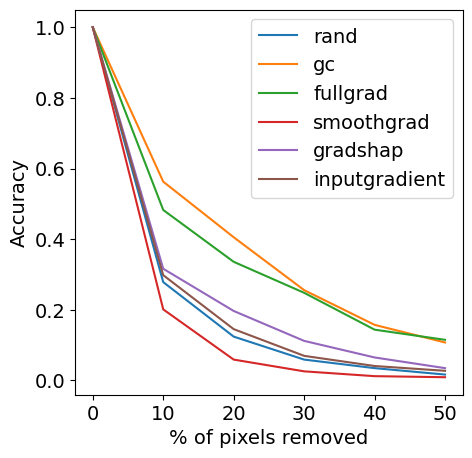}} \hspace{0.05cm} & 
 {\includegraphics[scale=0.2]{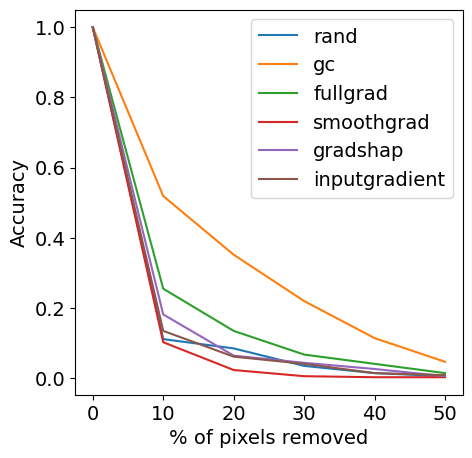}}\hspace{0.05cm}&
 {\includegraphics[scale=0.2]{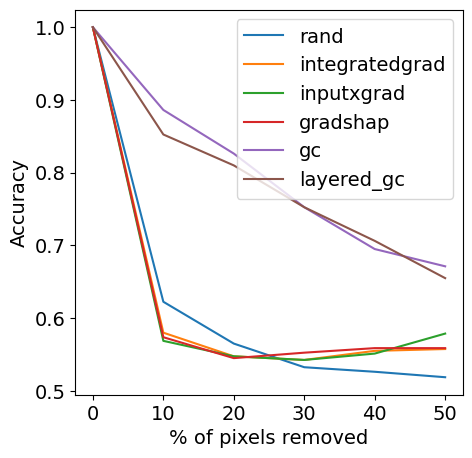}} & {\includegraphics[scale=0.2]{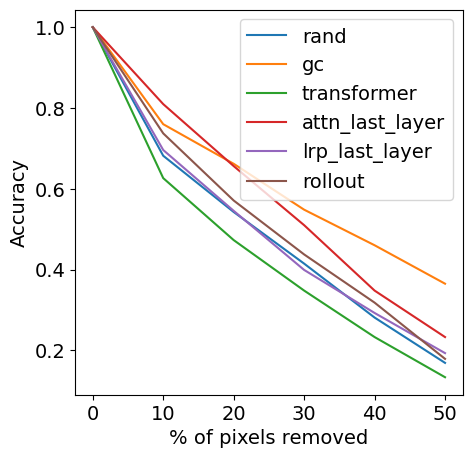}} \\
      \raisebox{25px}{\rotatebox{90}{Blur\cite{fong2017interpretable}}}
 &  
{\includegraphics[scale=0.2]{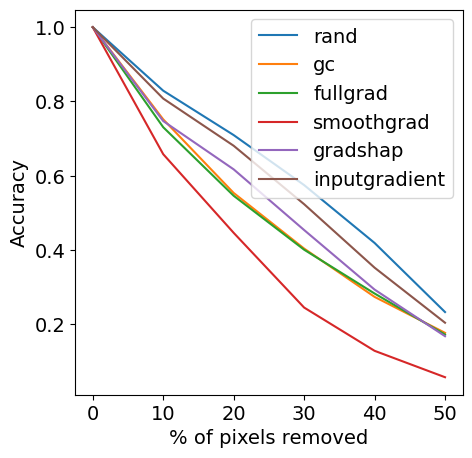}} \hspace{0.05cm} 
& {\includegraphics[scale=0.2]{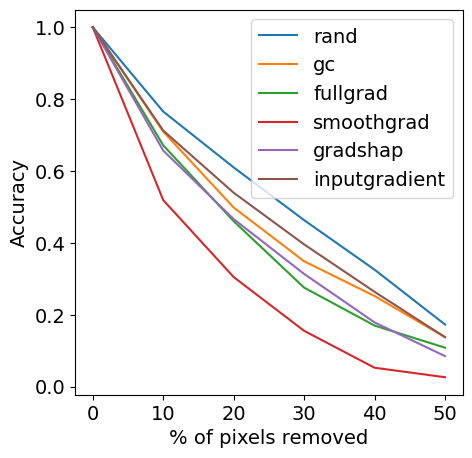}}\hspace{0.05cm} 
& {\includegraphics[scale=0.2]{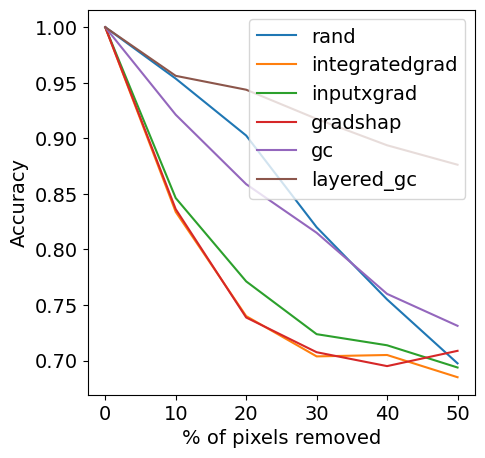}} & {\includegraphics[scale=0.2]{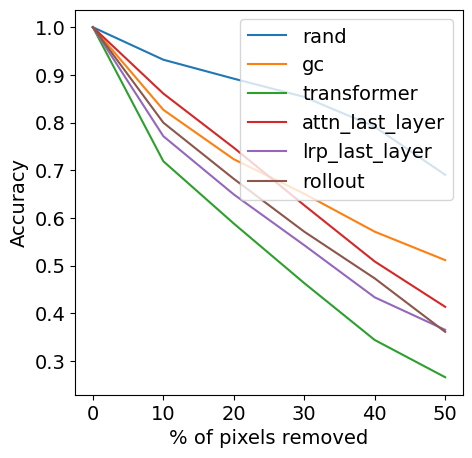}}\\
\raisebox{25px}{\rotatebox{90}{Mean\cite{hooker2019benchmark}}}
 &  
{\includegraphics[scale=0.2]{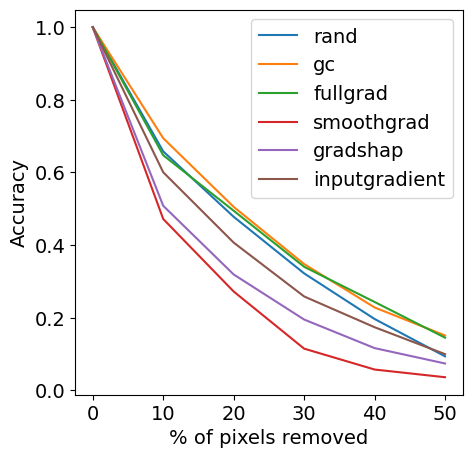}} \hspace{0.05cm}
& {\includegraphics[scale=0.2]{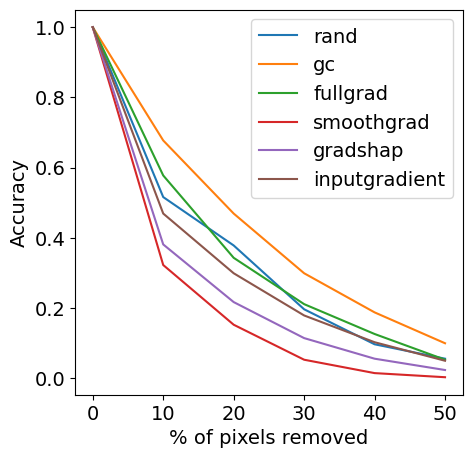}}\hspace{0.05cm} 
& {\includegraphics[scale=0.2]{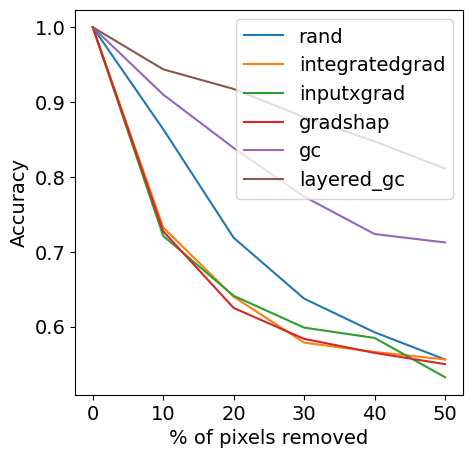}} & {\includegraphics[scale=0.2]{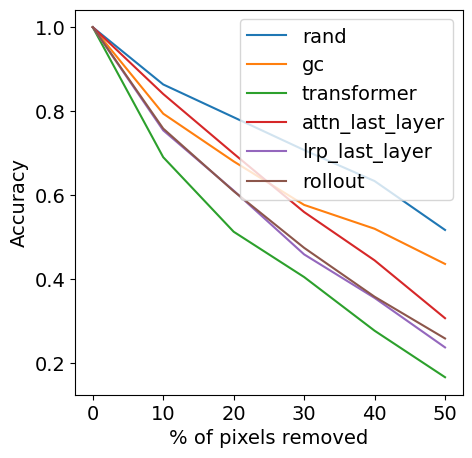}}\\
\raisebox{30px}{\rotatebox{90}{Ours}}
&  
{\includegraphics[scale=0.2]{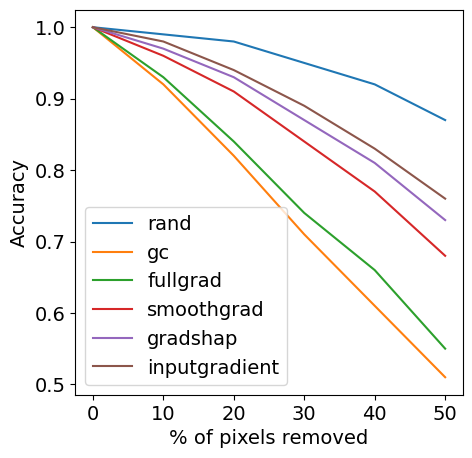}} \hspace{0.05cm} 
& {\includegraphics[scale=0.2]{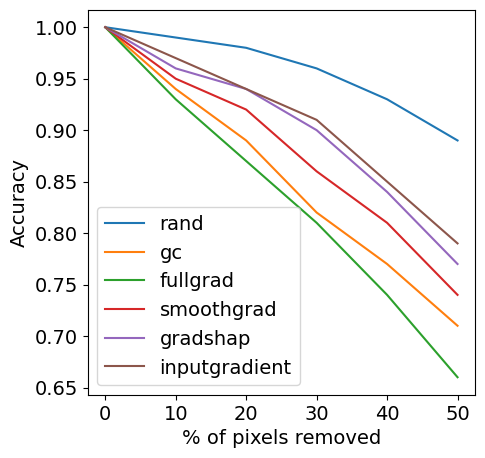}}\hspace{0.05cm} 
& {\includegraphics[scale=0.2]{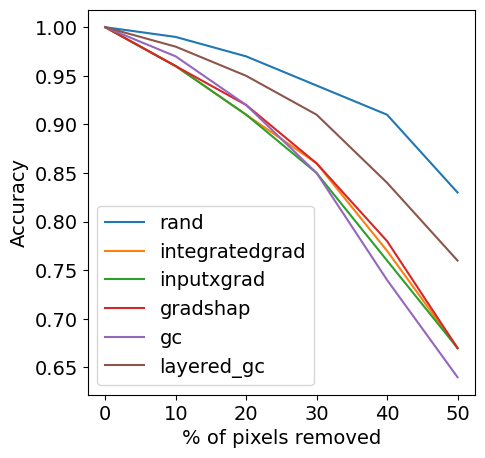}} & {\includegraphics[scale=0.2]{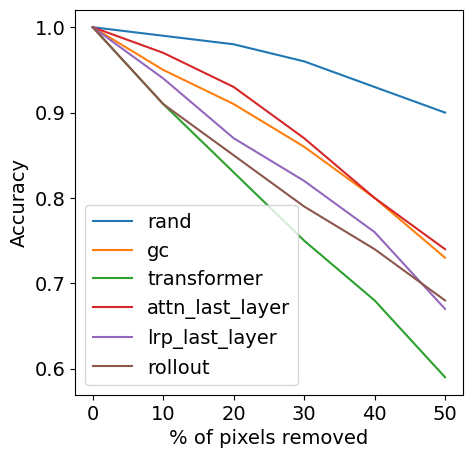}}\\
    \end{tabular}
    
    }
    \vspace{-8px}
    \caption{
    \textbf{Perturbation comparison against the leading baselines} with ResNet-50, VGG, AlexNet-based binary classifier, and ViT-B (please zoom in to view better). 
    % Perturbation tests with ResNet-50, AlexNet-based binary classifier, and ViT-B on the same set of images selected by step I of our method (see Sec. \ref{sec:method}). 
    For each model, we consider the most common explainability algorithms, in addition to a \emph{random} selection of pixels (see Sec \ref{sec:experimental_setup} for details).
    As can be observed, the baselines often struggle with separating random maps from actual relevance maps (\eg, delete for all models, blur for AlexNet, mean for all CNN variants) and appear to produce very similar results for all methods. Conversely, our method produces consistent ranking and meaningful distinction from the random baseline.
    }
    \label{fig:pertcompare}
    \vspace{-4.5px}
\end{figure*}

\begin{figure*}[t!]
        \begin{center}
             \includegraphics[width=1.05\linewidth]{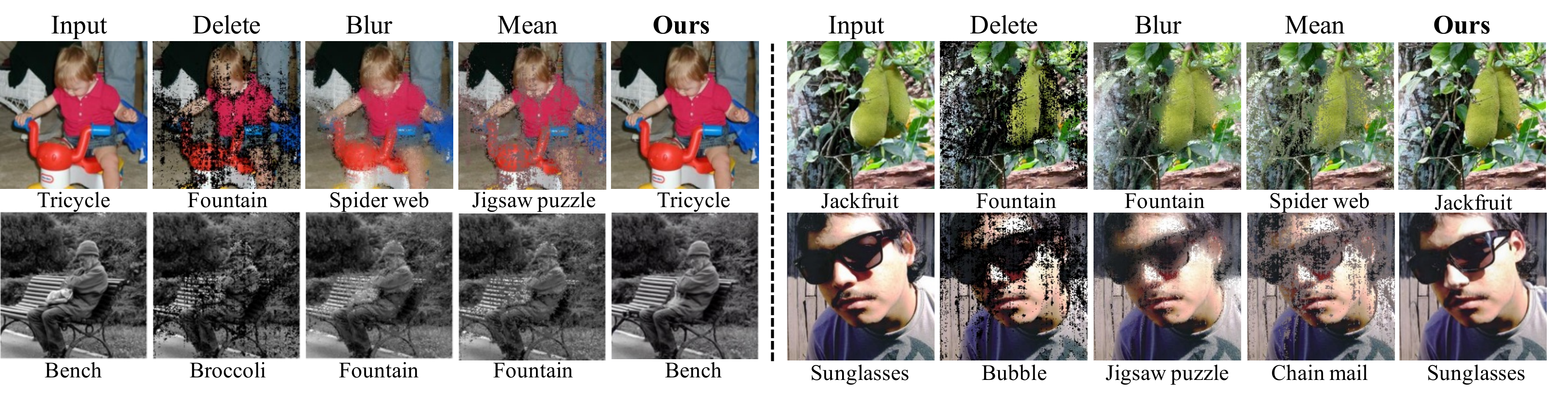}
        \end{center}
    \vspace{-16px}
\vspace{-4px}
    \caption{
    \textbf{Qualitative comparison for SmoothGrad.} According to all baselines, SmoothGrad is the leading method for CNN-based networks. We demonstrate examples where the baseline metrics indicate success for SmoothGrad (\ie, the prediction changed) while ours indicates failure (\ie, the prediction did not change to the top-2 class). The relevance maps produced by SmoothGrad often lead to OOD effects with the baselines (similar to Fig. \ref{fig:rand_mask}), thus their reliability for these results is questionable.}
    \label{fig:SmoothGrad}
    \vspace{-12px}
\end{figure*}

\begin{figure*}[t!]
        \begin{center}
             \includegraphics[width=\linewidth]{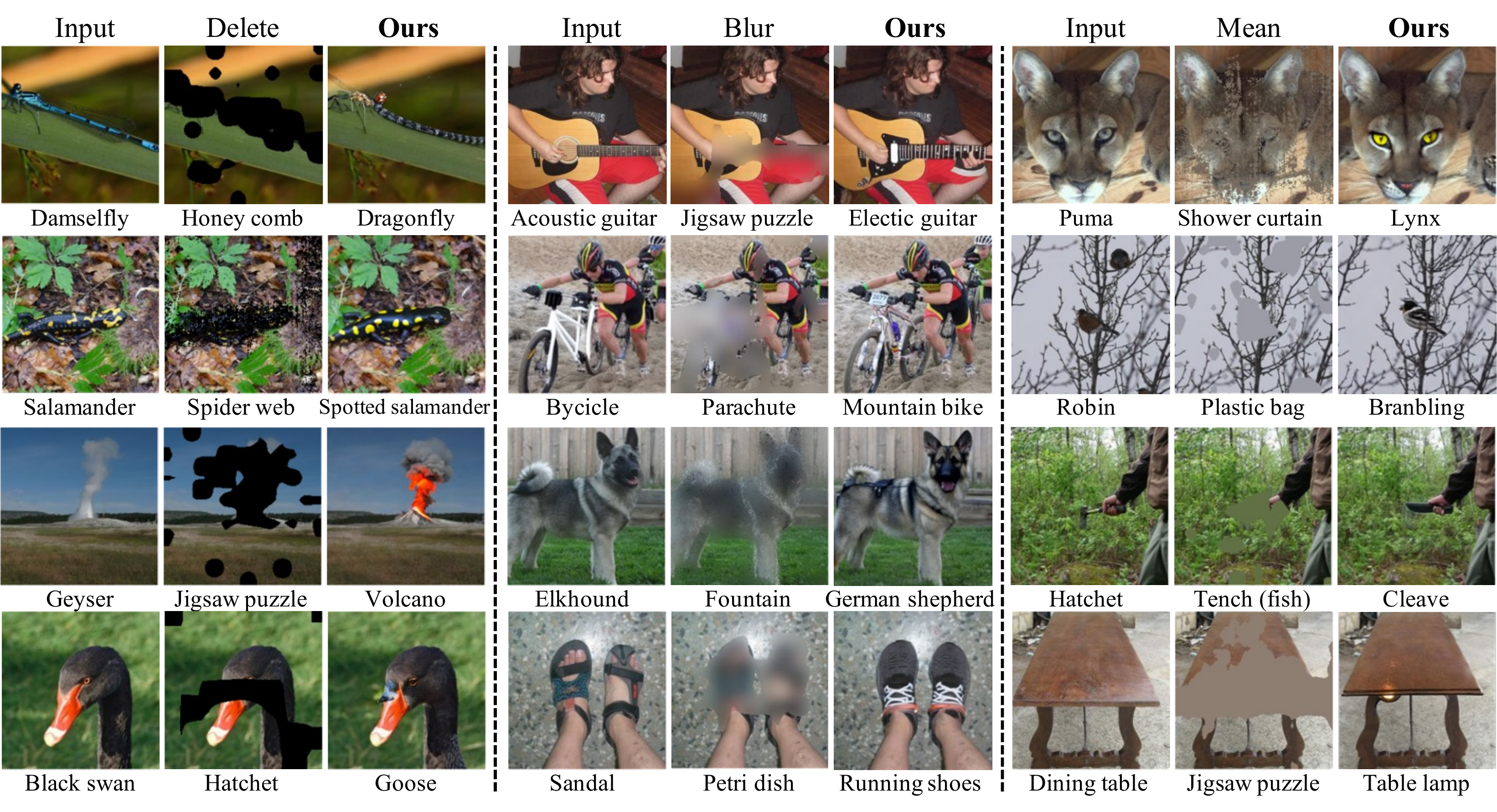}
        \end{center}
    \vspace{-18px}
    \caption{
    \textbf{Qualitative comparison of successful class changes against baselines.} We showcase image examples (Input) where both our method and the baselines induced a change in prediction (predictions indicated below each image). As seen, even when our method agrees with the baselines (\ie, the relevance map is faithful by all metrics), our method produces plausible pixel changes, while baselines cause OOD predictions.}
    \label{fig:qualitative}
    \vspace{-20px}
\end{figure*}

In this section, we demonstrate our method's advantage over the baselines in generating more natural results and producing a statistically significant separation between the different explainability methods. First, we qualitatively show the corresponding graphs for each of the evaluation methods (ours and the baselines) side by side for removing $p\in \{10\%, \dots, 50\%\}$ of the pixels (see Sec. \ref{sec:method}). As can be seen in Fig. \ref{fig:pertcompare}, our method is able to provide a meaningful ranking of the methods, where different explainability methods are \emph{distant from each other}, while also \emph{significantly separating the viable explanations from the random pixel selection baseline}. Conversely, the baseline evaluation metrics often produce graphs that are very similar for most of the attribution methods (see ResNet, VGG graphs), making it hard to tell whether a method's advantage is due to a statistical error or actual faithfulness to the classifier. The separation between different explainability algorithms is instrumental for the ability to empirically support the advantage of one explainability algorithm over the other, since the methods often produce \emph{entirely different results} (see supplementary materials for such examples). 

Additionally, the ranking of methods by the metrics often varies. For example, for all CNN variants, the baselines rank SmoothGrad as the top method, while our metric favors GradCAM or FullGrad. We hypothesize that this is due to SmoothGrad's relevance maps being scattered throughout the image, making them more susceptible to OOD effects, as shown in Fig. \ref{fig:rand_mask}. Fig. \ref{fig:SmoothGrad} qualitatively supports this claim, showing examples where the baselines indicated success with SmoothGrad (prediction changed), while our method showed the relevance was not faithful (prediction unchanged). These examples are similar to those in Fig. \ref{fig:rand_mask}, where relevance maps appear random or scattered, and the perturbed classification does not correlate with the image content.

Next, Fig. \ref{fig:qualitative} presents a qualitative comparison for cases where our method and the baselines \emph{agree}, \ie, both determine that the relevance map is faithful, as the perturbation successfully changed the prediction. Note that even in these cases, bizarre predictions are still observed when considering the baselines (\eg, jigsaw puzzle, fountain), while our method demonstrates actual plausible modifications of the class. This further indicates that even when our metric agrees with the baselines, it produces a much more coherent result, one that humans can understand \emph{why the metric deemed the relevance to be faithful}.

\subsection{Quantitative results}

\begin{table}[t]
\caption{
\textbf{Quantitative evaluation of sensitivity to OOD effects.} We evaluate the average distance from the random relevance baseline to estimate the impact of OOD inputs. A higher result indicates better separation.}
\vspace{-6px}
\centering
\begin{tabular}{@{}l@{~~} c@{~} c@{~} c@{~} c@{}}%
\toprule
% 1 - inpainting deletion blur mean
% -0.5729783037	0.1845771822	-0.5913758532	-0.5415976854

Method & ResNet & VGG & AlexNet & ViT\\
\midrule
 Delete \cite{hooker2019benchmark} & 0.26 &  0.18 & 0.32&  0.19 \\
 Blur \cite{fong2017interpretable}   & 0.43 & 0.42  & 0.20 & \textbf{0.59}  \\
 Mean \cite{hooker2019benchmark}   & 0.22 &  0.10 & 0.07 &  0.54  \\
 Ours & \textbf{0.49}  & \textbf{0.55}& \textbf{0.69}  & {0.57}\\
\bottomrule 
% \\[-0.3cm]
\end{tabular}
\label{tab:qualitative}
\vspace{-8px}
\end{table}

\begin{figure*}[t!]
\small
% \vspace{-10px}
\centering
\begin{tabular}{@{} c@{~~~~~} c@{}}%
{\includegraphics[scale=0.17]{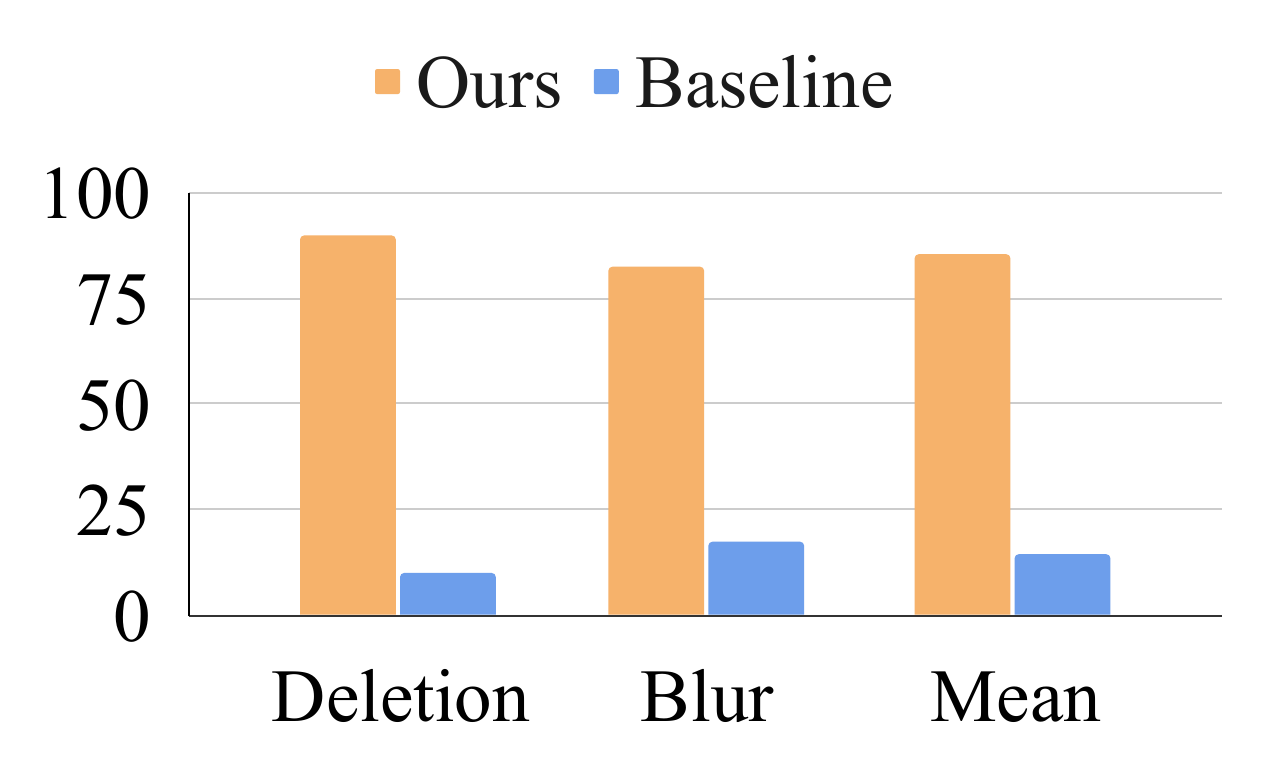}} & {\includegraphics[scale=0.17]{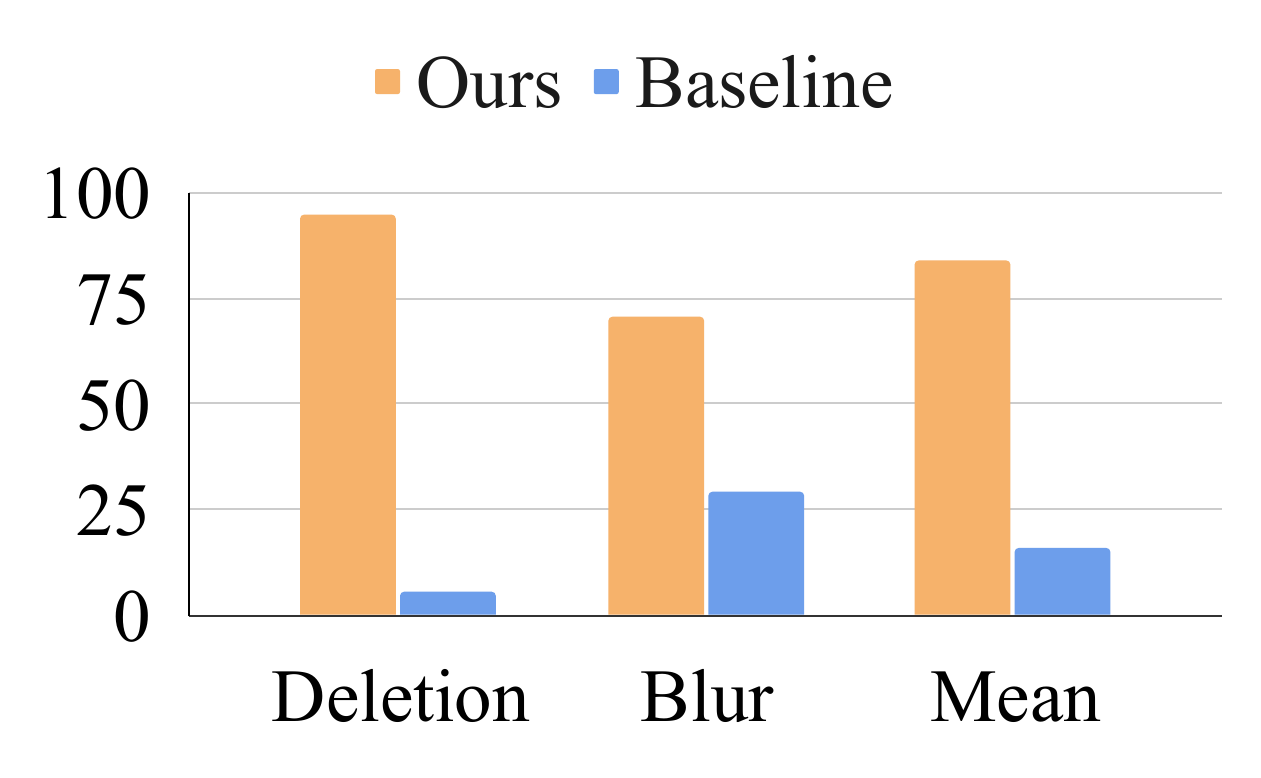}}   \\
(a) ResNet & (b) ViT 
% \\[-0.3cm]
\end{tabular}
\vspace{-6px}
\caption{
\textbf{User study evaluating the plausibility of our metric and baselines}. We randomly select cases where the best relevance maps differs for (a) ResNet and (b) ViT. Users select the most plausible map based on the input and prediction. The table shows the percentage favoring our method, with users overwhelmingly preferring it.}

\vspace{-18px}
\label{tab:user_study}
\end{figure*}

\paragraph{\textbf{Metrics.}} As mentioned, the main issue with existing methods is that the perturbed images are OOD, which causes prediction changes even for a random selection of pixels to remove. Therefore, we use the random pixel selection scores as a lower bound to estimate the impact of the OOD inputs on the method. For each evaluation method, we consider the overall score of each explainability method to be the area-under-the-curve (AUC) of the resulting graph (see graphs in Fig. \ref{fig:pertcompare}). This results in an aggregated score for all the explainability methods by each of the evaluation methods (ours and the baselines). Next, to compare the different evaluation metrics, we wish to estimate how well each metric separates the viable explainability algorithms from the OOD results of the random baselines. We calculate the average distance of the scores from the random lower bound as follows: Let $B$ denote the evaluation metric (ours or the baseline) that produces an AUC score for each attribution method ($\textbf{AUC}_B(E)$). We begin by normalizing the AUC results for $B$, to ensure that the different scales of the evaluation metrics do not impact our calculation, \ie: 
% \vspace{-5px}
\begingroup
\setlength{\abovedisplayskip}{5pt}
\setlength{\belowdisplayskip}{5pt}
\begin{equation}
    \forall E \in \{E_1, \dots, E_n\}: \textbf{AUC}_B(E) = \frac{\textbf{AUC}_B(E) - \min_{E}\textbf{AUC}_B(E)}{\max_{E}\textbf{AUC}_B(E) - \min_{E}\textbf{AUC}_B(E)}
\end{equation}
\endgroup
Next, we calculate the average distance between the viable algorithms and the random relevance baseline as follows:
\begingroup
\setlength{\abovedisplayskip}{5pt}
\setlength{\belowdisplayskip}{5pt}
\begin{equation}
    rand\_dist(B) =| \frac{1}{n}\sum_{E \in \{E_1, \dots, E_n\}} \textbf{AUC}_B(E) - \textbf{AUC}_B(rand)|,
\end{equation}
\endgroup
The comparison results are presented in Tab. \ref{tab:qualitative}. As can be seen, compared to the baselines, our method produces scores significantly more remote from the random lower bound for all CNN variants. Especially for the binary classifier and VGG, where the gaps are at least $37\%$ and $13\%$, respectively, in favor of our method. This shows our method's robustness to OOD inputs, rating viable attribution methods much higher than random pixel selection, with a greater ranking gap than the baselines. For ViT, our results align with pixel-blur and per-channel-mean, indicating that ViT is more robust to OOD inputs than CNNs. 

\paragraph{\textbf{User study.} }
As mentioned in Sec. \ref{sec:intro}, \emph{plausibility} is key in evaluating attribution methods, reflecting human interpretability of explanations. As shown in Fig. \ref{fig:pertcompare}, our ranking of explainability algorithms differs significantly from baselines, especially for CNNs. In addition to the improved faithfulness demonstrated above, we verify its correlation with human preferences, ensuring greater \emph{plausibility}.

Hence, we select ResNet (CNN) and ViT (Transformer) as representatives and sample 10 images per baseline (delete, blur, mean), where our metric and the baseline favor different attributions. We use the Two-Alternative Forced Choice (2AFC) protocol, where users compare relevance maps from both methods and select the one best aligned with the input and classification. Fig. \ref{tab:user_study} shows results from 22 participants and 656 responses. When our metric disagrees with baselines, users overwhelmingly prefer our top-ranked attribution, demonstrating improved \emph{faithfulness} and \emph{plausibility}. This shows the properties align and can be optimized together. Fleiss' Kappa (mean 0.79, std 0.18) indicates strong inter-reviewer agreement, suggesting minimal subjective bias.

\paragraph{\textbf{Ablation study.} }

\begin{figure*}[t!]
    \centering
    \setlength{\tabcolsep}{0.5pt}
    \addtolength{\belowcaptionskip}{-8pt}
    {\small
    \begin{tabular}{l@{~} c c c c}
    &
        (a) & (b) & (c) & (d)\\
        \raisebox{15px}{\begin{turn}{90}~~~ ResNet \end{turn}} 
 & {\includegraphics[scale=0.2]{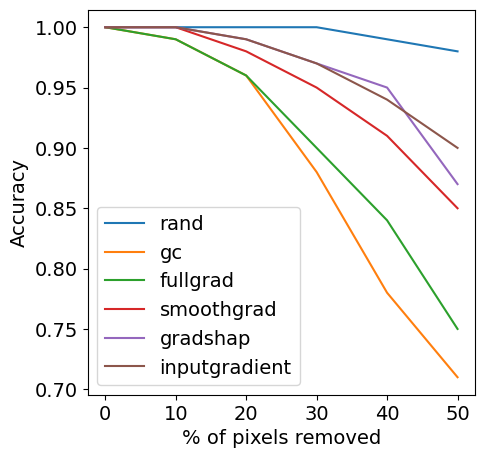}} \hspace{0.05cm} & 
 {\includegraphics[scale=0.2]{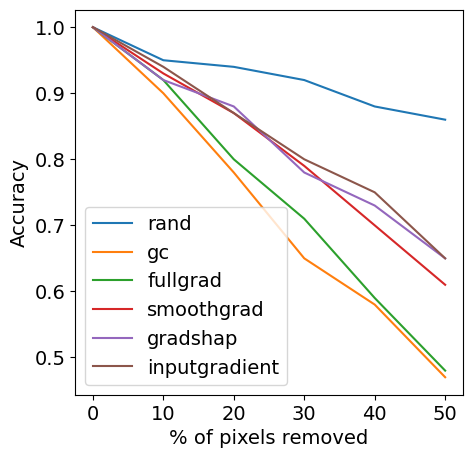}}\hspace{0.05cm}&
 {\includegraphics[scale=0.2]{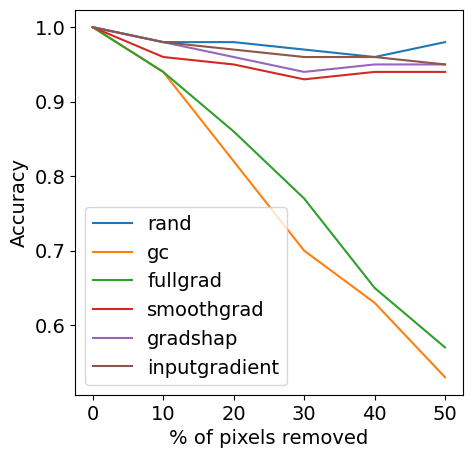}} & {\includegraphics[scale=0.2]{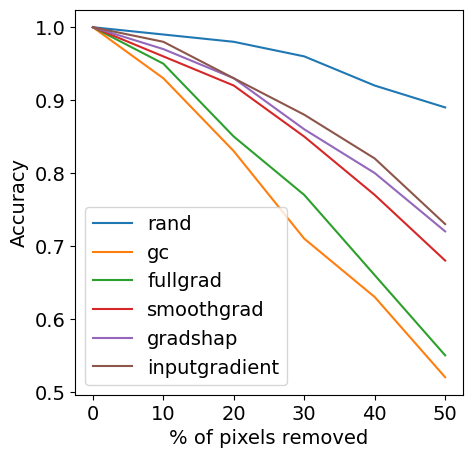}} \\
      \raisebox{35px}{\rotatebox{90}{ViT}}
 &  
{\includegraphics[scale=0.2]{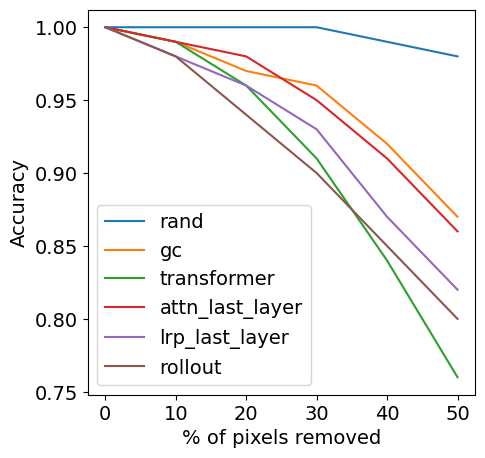}} \hspace{0.05cm} 
& {\includegraphics[scale=0.2]{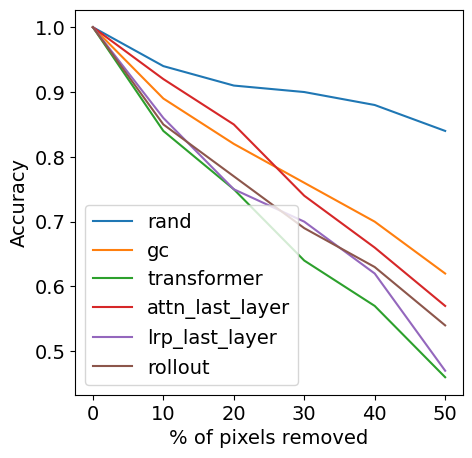}}\hspace{0.05cm} 
& {\includegraphics[scale=0.2]{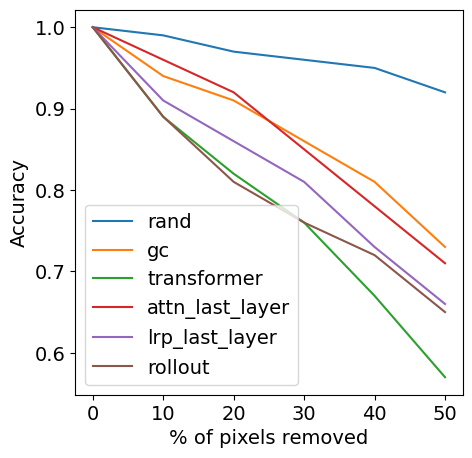}} & {\includegraphics[scale=0.2]{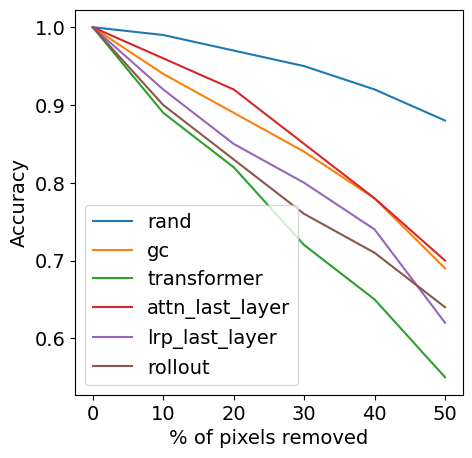}}\\
    \end{tabular}
    
    }
    \vspace{-8px}
    \caption{
    \textbf{Ablation study.} 
    We compare our full method with different variants of it on ResNet and ViT; (a) top-10- using the top-10 class instead of the top-2, (b) no weighting- results without applying our weighting function, (c) inpaint pixelwise- performing pixel-wise inpainting instead of patch-base inpainting, (d) full method.
    }
    \label{fig:ablation}
    \vspace{-8px}
\end{figure*}

In Fig. \ref{fig:ablation}, we consider three variants of our method to estimate the impact of each design choice. (1) \emph{Top 10 inpainting}: Instead of inpainting to the top-2 class, we use the top-10 class. While top-2 classes share more semantic features, improving SD inpainting, inpainting to top-10 slightly reduces stability. Yet, the random baseline remains distinct, and top methods stay consistent, demonstrating robustness. (2)  \emph{no weighting}: Removing the weighting function (Alg. \ref{alg:method}, line 10) makes attribution methods cluster more, whereas with it, they are better distinguished. Clear differentiation is crucial for a meaningful metric, as methods often yield entirely different relevance maps for the same input (see supplementary materials). (3)\emph{Pixel-wise inpainting}: We remove patch-wise downsampling (Alg. \ref{alg:method}, line 3), forcing pixel-level inpainting, which is out-of-distribution for SD. Results confirm that patch-wise inpainting is crucial, especially for CNN-based models.

\section{Discussion and Limitations}
As shown in our experiments, our method is a viable, stable alternative to perturbation-based evaluations. However, there are some limitations to consider. First, a limitation shared among perturbation methods is that they are only applied to the top-1 classification, not to all target classes. While some explainability methods, e.g.,~\cite{selvaraju2017grad,Chefer_2021_CVPR} can provide explanations for any class (\ie, \emph{class-specific explainability methods}), perturbation-based metrics only measure the faithfulness of the relevance map for the top-1 prediction of the classifier. A generalization to additional classes could provide deeper insight into the attribution method and indicate whether the provided class-specific signal is meaningful. Additionally, since we rely on an inpainting engine, we are limited to test images for which where the alternative (top-2) class is recognized by that engine (see step I of our method in Sec. \ref{sec:method}). One solution is to fine-tune the inpainting engine for all classes of a given classifier, but that is beyond our scope. Finally, inpainting is time-consuming, making evaluation on very big datasets harder, we believe progress in diffusion model inference time \cite{lcmlora2023} will soon alleviate this hurdle, allowing our metric to be used for any data size.

\section{Conclusion}
As deep neural networks (DNNs) permeate everyday applications, the need for explainability has become paramount. Attribution methods offer an intuitive means of explanation serving diverse purposes, from enhancing classifier robustness \cite{chefer2022robustvit} to guiding generative models \cite{chefer2023attendandexcite}. However, current evaluation metrics for attribution methods suffer from severe out-of-distribution (OOD) issues, rendering them unreliable. To address this challenge, we propose a novel metric, leveraging the potent inpainting capabilities of Stable Diffusion (SD). Our metric excels in both faithfulness and plausibility, crucial elements of an effective explanation. As generative models continue to advance, we believe that the fusion of explainability and generative capabilities is poised to play an increasingly prominent role in enhancing research in both fields.

\section{Acknowledgements}
This work was supported by the Tel Aviv University Center for AI and Data Science (TAD).

\bibliographystyle{splncs04}
\bibliography{scia}

\begin{thebibliography}{10}
\providecommand{\url}[1]{\texttt{#1}}
\providecommand{\urlprefix}{URL }
\providecommand{\doi}[1]{https://doi.org/#1}

\bibitem{abnar2020quantifying}
Abnar, S., et~al.: Quantifying attention flow in transformers. In: ACL (2020)

\bibitem{adebayo2018sanity}
Adebayo, J., et~al.: Sanity checks for saliency maps. NeurIPS  (2018)

\bibitem{agarwal2020explaining}
Agarwal, C., et~al.: Explaining image classifiers by removing input features using generative models. In: ACCV (2020)

\bibitem{bach2015pixel}
Bach, S., et~al.: On pixel-wise explanations for non-linear classifier decisions by layer-wise relevance propagation. PloS one  (2015)

\bibitem{binder2016layer}
Binder, A., et~al.: Layer-wise relevance propagation for neural networks with local renormalization layers. In: ICANN (2016)

\bibitem{Buolamwini2018GenderSI}
Buolamwini, J., et~al.: Gender shades: Intersectional accuracy disparities in commercial gender classification. In: FAT (2018)

\bibitem{chang2018explaining}
Chang, C.H., et~al.: Explaining image classifiers by adaptive dropout and generative in-filling. arXiv:1807.08024  (2018)

\bibitem{Chefer_2021_ICCV}
Chefer, H., et~al.: Generic attention-model explainability for interpreting bi-modal and encoder-decoder transformers. In: Proceedings of the IEEE/CVF ICCV (2021)

\bibitem{Chefer_2021_CVPR}
Chefer, H., et~al.: Transformer interpretability beyond attention visualization. In: Proceedings of the IEEE/CVF CVPR (2021)

\bibitem{chefer2022robustvit}
Chefer, H., et~al.: Optimizing relevance maps of vision transformers improves robustness. Advances in NeurIPS  (2022)

\bibitem{chefer2023attendandexcite}
Chefer, H., et~al.: Attend-and-excite: Attention-based semantic guidance for text-to-image diffusion models (2023)

\bibitem{chen2018lshapley}
Chen, J., et~al.: L-shapley and c-shapley: Efficient model interpretation for structured data. In: ICLR (2019)

\bibitem{dabkowski2017real}
Dabkowski, P., Gal, Y.: Real time image saliency for black box classifiers. In: Advances in NeurIPS (2017)

\bibitem{Dabkowski2017RealTI}
Dabkowski, P., Gal, Y.: Real time image saliency for black box classifiers. In: Neural Information Processing Systems (2017), \url{https://api.semanticscholar.org/CorpusID:41766449}

\bibitem{DeYoung2019ERASERAB}
DeYoung, J., et~al.: Eraser: A benchmark to evaluate rationalized nlp models. In: ACL (2019)

\bibitem{Dhamdhere2018HowII}
Dhamdhere, K., et~al.: How important is a neuron? ArXiv  \textbf{abs/1805.12233} (2018)

\bibitem{dosovitskiy2020image}
Dosovitskiy, A., et~al.: An image is worth 16x16 words: Transformers for image recognition at scale. arXiv:2010.11929  (2020)

\bibitem{erhan2009visualizing}
Erhan, D., et~al.: Visualizing higher-layer features of a deep network. UdeM  (2009)

\bibitem{Fel2021WhatIC}
Fel, T., et~al.: What i cannot predict, i do not understand: A human-centered evaluation framework for explainability methods. ArXiv  \textbf{abs/2112.04417} (2021)

\bibitem{fong2019understanding}
Fong, R., et~al.: Understanding deep networks via extremal perturbations and smooth masks. In: ICCV (2019)

\bibitem{fong2017interpretable}
Fong, R.C., Vedaldi, A.: Interpretable explanations of black boxes by meaningful perturbation. In: ICCV (2017)

\bibitem{Fuster2017PredictablyUT}
Fuster, A., et~al.: Predictably unequal? the effects of machine learning on credit markets. Regulation of Financial Institutions eJournal  (2017)

\bibitem{Ghorbani2019TowardsAC}
Ghorbani, A., Wexler, J., Zou, J.Y., Kim, B.: Towards automatic concept-based explanations. In: NeurIPS (2019)

\bibitem{Goyal2019CounterfactualVE}
Goyal, Y., et~al.: Counterfactual visual explanations. In: ICML (2019)

\bibitem{guillaumin2014imagenet}
Guillaumin, M., et~al.: Imagenet auto-annotation with segmentation propagation. IJCV  (2014)

\bibitem{he2016deep}
He, K., Zhang, X., Ren, S., Sun, J.: Deep residual learning for image recognition. In: Proceedings of the IEEE Conference on Computer Vision and Pattern Recognition. pp. 770--778 (2016)

\bibitem{He2015DeepRL}
He, K., et~al.: Deep residual learning for image recognition. CVPR  (2015)

\bibitem{Hendrycks2019BenchmarkingNN}
Hendrycks, D., Dietterich, T.G.: Benchmarking neural network robustness to common corruptions and perturbations. ArXiv  \textbf{abs/1903.12261} (2019)

\bibitem{Hendrycks2019NaturalAE}
Hendrycks, D., et~al.: Natural adversarial examples. CVPR  (2019)

\bibitem{hooker2019benchmark}
Hooker, S., et~al.: A benchmark for interpretability methods in deep neural networks. In: Advances in NeurIPS (2019)

\bibitem{jacovi-goldberg-2020-towards}
Jacovi, A., Goldberg, Y.: Towards faithfully interpretable {NLP} systems: How should we define and evaluate faithfulness? In: ACL (2020)

\bibitem{Kim2017InterpretabilityBF}
Kim, B., et~al.: Interpretability beyond feature attribution: Quantitative testing with concept activation vectors (tcav). In: ICML (2017)

\bibitem{krizhevsky2012imagenet}
Krizhevsky, A., et~al.: Imagenet classification with deep convolutional neural networks. In: NeurIPS (2012)

\bibitem{Li2016UnderstandingNN}
Li, J., et~al.: Understanding neural networks through representation erasure. ArXiv  \textbf{abs/1612.08220} (2016)

\bibitem{Liu2022RethinkingAE}
Liu, Y., et~al.: Rethinking attention-model explainability through faithfulness violation test. ArXiv  \textbf{abs/2201.12114} (2022)

\bibitem{lundberg2017unified}
Lundberg, S., et~al.: A unified approach to interpreting model predictions. In: NeurIPS (2017)

\bibitem{lcmlora2023}
Luo, S., et~al.: Lcm-lora: A universal stable-diffusion acceleration module (2023)

\bibitem{mahendran2016visualizing}
Mahendran, A., Vedaldi, A.: Visualizing deep convolutional neural networks using natural pre-images. IJCV  (2016)

\bibitem{montavon2017explaining}
Montavon, G., et~al.: Explaining nonlinear classification decisions with deep taylor decomposition. Pattern Recognition  (2017)

\bibitem{nichol2021glide}
Nichol, A., et~al.: Glide: Towards photorealistic image generation and editing with text-guided diffusion models. arXiv:2112.10741  (2021)

\bibitem{Ramesh2022Hierarchical}
Ramesh, A., et~al.: Hierarchical text-conditional image generation with clip latents. ArXiv  (2022), \url{https://arxiv.org/abs/2204.06125}

\bibitem{Rombach2021HighResolutionIS}
Rombach, R., et~al.: High-resolution image synthesis with latent diffusion models. IEEE/CVF CVPR  (2021)

\bibitem{russakovsky2015ImageNet}
Russakovsky, O., et~al.: Imagenet large scale visual recognition challenge. IJCV  (2015)

\bibitem{Saharia2022PhotorealisticTD}
Saharia, C., et~al.: Photorealistic text-to-image diffusion models with deep language understanding. ArXiv  \textbf{abs/2205.11487} (2022)

\bibitem{2303.17155}
Schwartz, I., et~al.: Discriminative class tokens for text-to-image diffusion models (2023)

\bibitem{selvaraju2017grad}
Selvaraju, R.R., et~al.: Grad-cam: Visual explanations from deep networks via gradient-based localization. In: ICCV (2017)

\bibitem{shrikumar2017learning}
Shrikumar, A., Greenside, P., Kundaje, A.: Learning important features through propagating activation differences. In: ICML (2017)

\bibitem{Simonyan2014VeryDC}
Simonyan, K., Zisserman, A.: Very deep convolutional networks for large-scale image recognition. CoRR  \textbf{abs/1409.1556} (2014)

\bibitem{simonyan2013deep}
Simonyan, K., et~al.: Deep inside convolutional networks: Visualising image classification models and saliency maps. arXiv:1312.6034  (2013)

\bibitem{smilkov2017SmoothGrad}
Smilkov, D., et~al.: Smoothgrad: removing noise by adding noise. arXiv:1706.03825  (2017)

\bibitem{srinivas2019full}
Srinivas, S., Fleuret, F.: Full-gradient representation for neural network visualization. In: NeurIPS (2019)

\bibitem{Sundararajan2017AxiomaticAF}
Sundararajan, M., et~al.: Axiomatic attribution for deep networks. In: ICML (2017)

\bibitem{voita2019analyzing}
Voita, E., et~al.: Analyzing multi-head self-attention: Specialized heads do the heavy lifting, the rest can be pruned. In: ACL (2019)

\bibitem{zeiler2014visualizing}
Zeiler, M.D., et~al.: Visualizing and understanding convolutional networks. In: ECCV (2014)

\bibitem{zhang2018top}
Zhang, J., et~al.: Top-down neural attention by excitation backprop. IJCV  (2018)

\bibitem{zhou2016learning}
Zhou, B., et~al.: Learning deep features for discriminative localization. In: Proceedings of CVPR (2016)

\end{thebibliography}

\appendix
\section{Implementation Details}

\noindent\textbf{Data Construction}

We create dataset $S$ with all images recognized by the SD model, following these steps: (1) For each class $t\in {1,\dots, T}$, we generate 20 images $i^t_1,\dots, i^t_{20}$ using the prompt "{class-name}". (2) We classify these images: $\forall{t\in 1,\dots, T}: C(i^t_1),\dots, C(i^t_{20})$ (3) We add class $t$ to $S$ if at least 50\% of its images are classified as $t$. This results in $|S|=661$ classes for ResNet and VGG, $|S|=704$ classes for ViT, and all classes for the binary classifier. Finally, we sample an image per class as in Step 2 of our method.

\noindent\textbf{Inpainting details}
We utilized the public version stable-diffusion-inpaiting~{\cite{Rombach_2022_CVPR}} from Hugging Face, initialized with the Stable-Diffusion-v-1-2 weights, guidance scale parameter was set to 7.5. We used NVIDIA GeForce RTX 2080 Ti GPU with 11GB of memory. Images and their masks were resized to $512 \times 512$ for inpainting and resized back to $224 \times 224$ before classification.

% \section{Variance in Attribution Methods}
% \label{sec:variance}
% \begin{figure}[t]
%     \centering
%     \addtolength{\belowcaptionskip}{-8pt}
%     \begin{subfigure}{0.49\linewidth}
%         \centering
%         \includegraphics[width=\linewidth]{images/different_attribution/resnet.pdf}
%         \caption{ResNet Attribution}
%     \end{subfigure}
%     \hfill
%     \begin{subfigure}{0.49\linewidth}
%         \centering
%         \includegraphics[width=\linewidth]{images/different_attribution/ViT.pdf}
%         \caption{ViT Attribution}
%     \end{subfigure}
%     \vspace{-1pt}
%     \caption{Relevance maps from different attribution methods for the same input using ResNet (left) and ViT (right), highlighting the need for a metric.}
%     \label{fig:resnet_heatmaps}
%     % \vspace{-8pt}
% \end{figure}

\section{Variance in Attribution Methods}
\label{sec:variance}
\begin{figure}[ht] % Changed [t] to [ht] for better placement possibility
    \centering
    \addtolength{\belowcaptionskip}{-8pt} % Adjusts space below the main caption

    \includegraphics[width=0.49\linewidth]{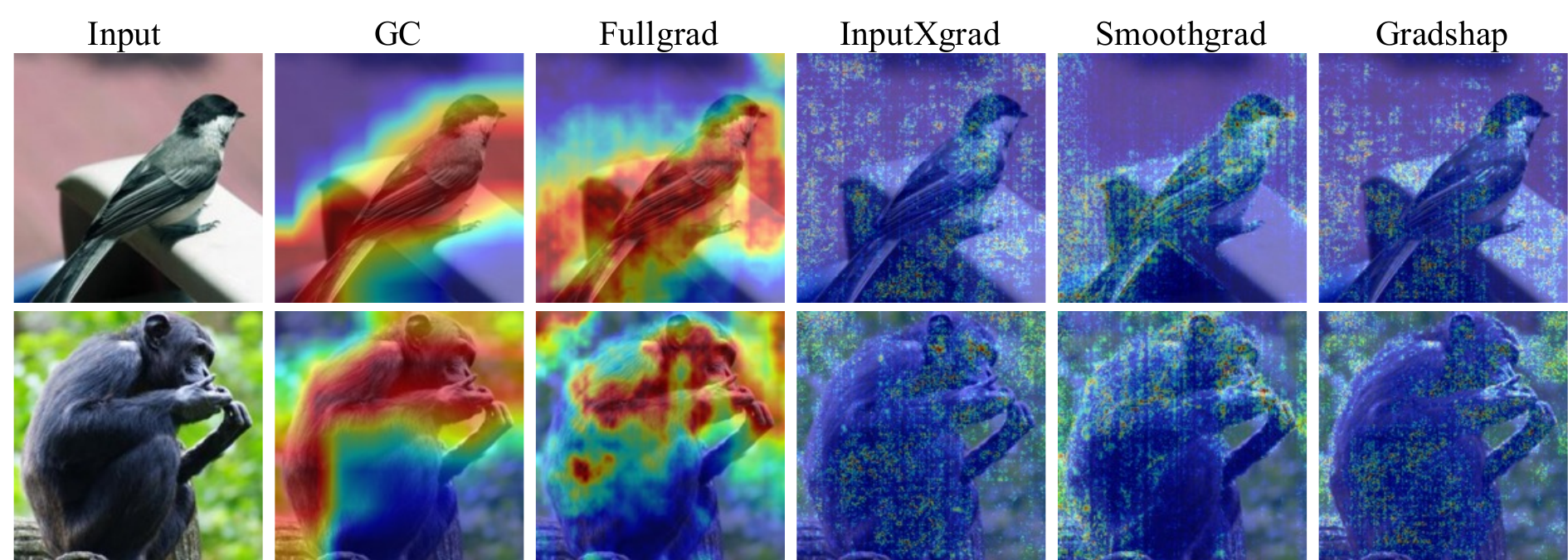} % Replace with your image path
    \hfill % Pushes the images apart to fill the line width
    \includegraphics[width=0.49\linewidth]{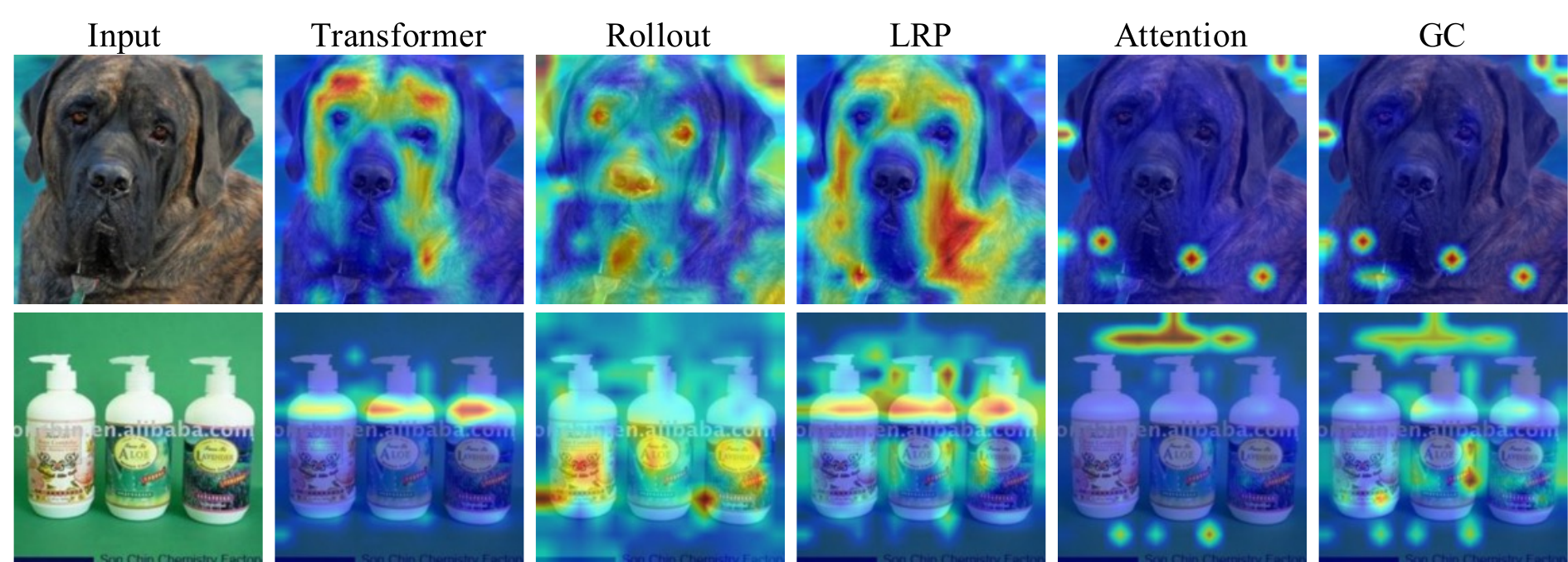} % Replace with your image path

    \caption{Relevance maps from different attribution methods for the same input using ResNet (left) and ViT (right), highlighting the need for a metric.}
    \label{fig:resnet_heatmaps}
    % \vspace{-8pt} % Removed this \vspace, adjust if needed
\end{figure}

As mentioned in the main paper, different attribution methods yield very different relevance maps. To empirically substantiate this point, Figure \ref{fig:resnet_heatmaps} illustrates this with examples from ResNet and ViT, showing that each method highlights different pixels for the same input. 

\section{Additional baselines}
\begin{figure*}[t!]
    \centering
    \setlength{\tabcolsep}{0.3pt}
    \addtolength{\belowcaptionskip}{-8pt}
    {\tiny
    \begin{tabular}{l c c c c}
    &
        ResNet \cite{he2016deep} & VGG \cite{simonyan2013deep} & AlexNet (binary) \cite{krizhevsky2012imagenet} & ViT \cite{dosovitskiy2020image}\\
        \raisebox{0.01px}{\begin{turn}{90}~~~ Violations \cite{Liu2022RethinkingAE} \end{turn}} 
 & {\includegraphics[scale=0.13]{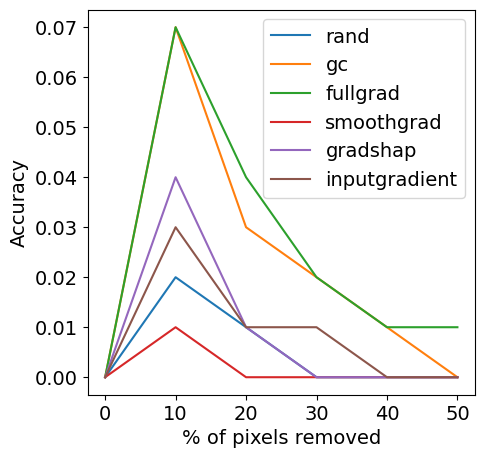}} \hspace{0.03cm} & 
 {\includegraphics[scale=0.13]{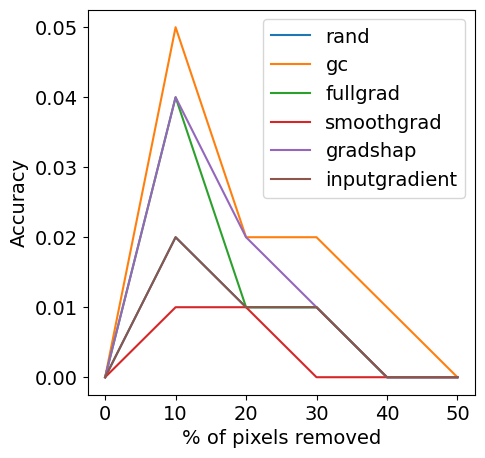}}\hspace{0.03cm}&
 {\includegraphics[scale=0.13]{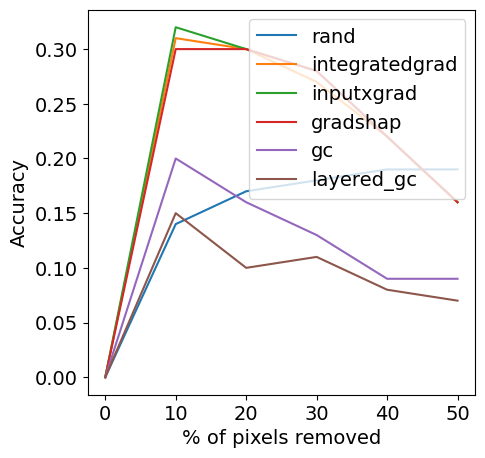}}\hspace{0.03cm}  & {\includegraphics[scale=0.13]{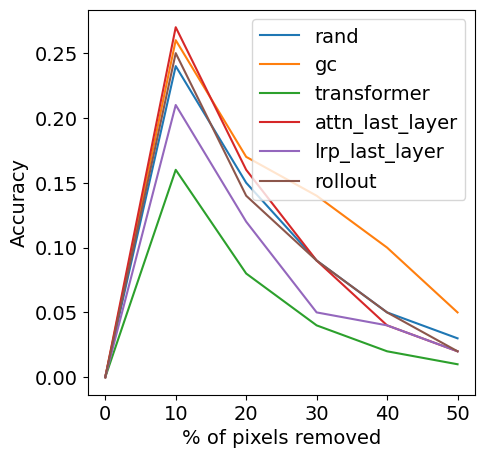}}\hspace{0.03cm}  \\
      \raisebox{8px}{\rotatebox{90}{Saliency\cite{Dabkowski2017RealTI}}}
 &  
{\includegraphics[scale=0.13]{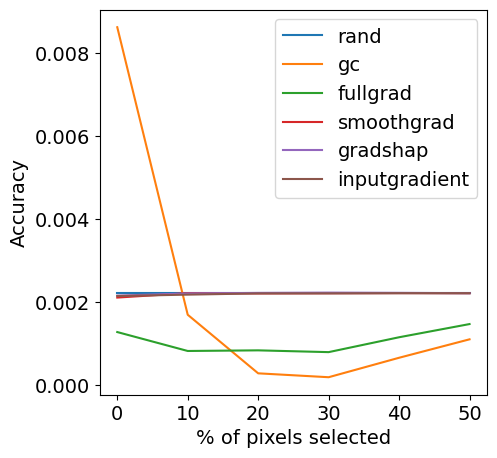}} \hspace{0.03cm} 
& {\includegraphics[scale=0.13]{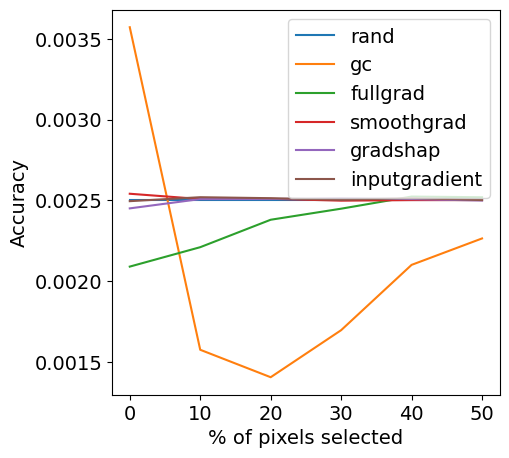}}\hspace{0.03cm} 
& {\includegraphics[scale=0.13]{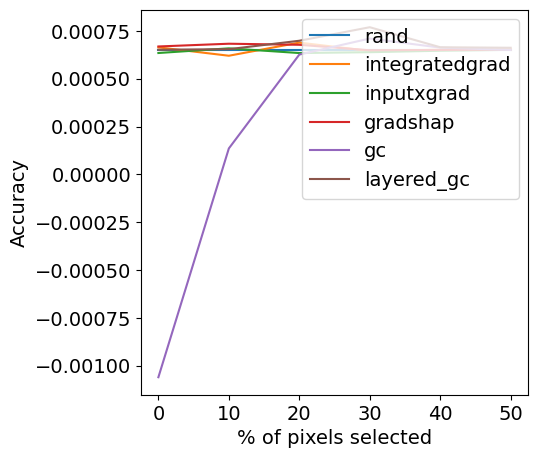}}\hspace{0.03cm}  & {\includegraphics[scale=0.13]{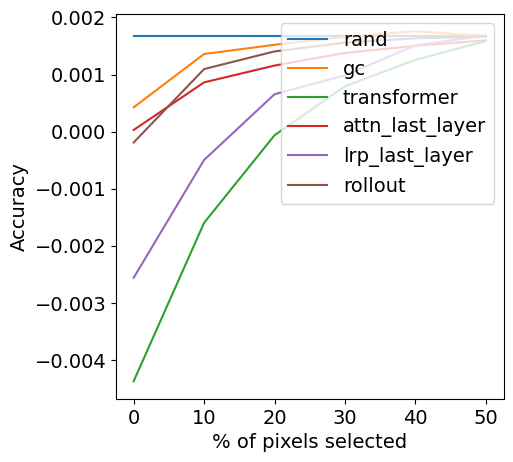}}\hspace{0.03cm} \\
\raisebox{20px}{\rotatebox{90}{Ours}}
 &  
{\includegraphics[scale=0.13]{images/pertcompare_new/resnet_positive_inpaint.png}} \hspace{0.03cm} 
& {\includegraphics[scale=0.13]{images/pertcompare_new/vgg_positive_inpaint.png}}\hspace{0.03cm} 
& {\includegraphics[scale=0.13]{images/pertcompare_new/binary_positive_inpaint.png}}\hspace{0.03cm}  & {\includegraphics[scale=0.13]{images/pertcompare_new/vit_positive_inpaint.png}}\hspace{0.03cm} \\
    \end{tabular}
    
    }
    \vspace{-8px}
    \caption{
    \textbf{Perturbation comparison against additional baselines} 
    % with ResNet-50, VGG, AlexNet-based binary classifier, and ViT-B
    % Perturbation tests with ResNet-50, AlexNet-based binary classifier, and ViT-B on the same set of images selected by step I of our method (see Sec. \ref{sec:method}). 
    We consider most common explainability methods and a \emph{random} pixel selection. These baselines often fail to distinguish relevance maps from random ones (\cite{Liu2022RethinkingAE}) and yield similar results across methods (\cite{Dabkowski2017RealTI}). 
    % In contrast, our method ensures consistent ranking and clear separation from the random baseline.
    }
    \label{fig:pertcompare_additoinal}
    \vspace{-8px}
\end{figure*}

We compare our methods to additional baselines. The first is faithfulness violation \cite{Liu2022RethinkingAE} which checks that the deletion of the relevant pixels decreases the confidence of the predicted class. The second baseline is a saliency-based evaluation \cite{Dabkowski2017RealTI} where for each percentage of perturbation pixels ($10\%, \dots, 50\%$), one first extracts the smallest rectangle patch that contains the top pixels, then applies the classifier to that patch to test whether the prediction remains the same.

Fig. \ref{fig:pertcompare_additoinal} shows that the violation method fails to separate the random baseline from real attribution methods, indicating that, like the baseline perturbation metrics, it is susceptible to OOD modifications. Additionally, the saliency-based metric often produces nearly indistinguishable outputs for various attribution methods (for all classifiers except ViT). Distinguishing these is crucial for evaluating different methods. Moreover, the random baseline cannot be clearly separated from valid attribution methods for all models except ViT.

\end{document}